\documentclass{article}
\pdfoutput=1 
\usepackage[12pt]{extsizes}
\usepackage[utf8]{inputenc}
\usepackage{amsmath}
\usepackage{amssymb}
\usepackage{booktabs}

\usepackage[square,sort,comma,numbers]{natbib}
\usepackage[pdftex]{graphicx}
\usepackage[dvipsnames]{xcolor}
\usepackage[hidelinks]{hyperref}
\hypersetup{
    colorlinks = true,
    linkcolor={blue!50!black},
    citecolor={blue!50!black},
    urlcolor={blue!50!black}
}
\usepackage{authblk}
\usepackage{geometry}
\usepackage{parskip} 
\usepackage[inline]{enumitem}

\usepackage{booktabs}  
\usepackage{amsfonts}
\usepackage{mathtools}
\usepackage{url}
\usepackage{enumitem}
\usepackage{authblk}
\usepackage{comment}
\usepackage{mdframed}
\usepackage{wrapfig}
\usepackage{floatrow}
\usepackage{pifont}

\newcommand{\hide}[1]{}
\usepackage{ntheorem}

\theoremseparator{:} 

\newmdtheoremenv[%
  backgroundcolor=white,
  linecolor=blue!60!black,
  linewidth=2pt,
  topline=true,
  rightline=false,
  skipabove=10pt,
  skipbelow=10pt,
  leftline=false]{ourexample}{Application}
  
\newmdtheoremenv[%
  backgroundcolor=gray!20,
  linecolor=red!60!black,
  linewidth=2pt,
  topline=false,
  rightline=false,
  skipabove=10pt,
  skipbelow=10pt,
  leftline=false]{ourbox}{Formulation}
  
\newmdtheoremenv[%
  backgroundcolor=gray!20,
  linecolor=red!60!black,
  linewidth=2pt,
  topline=false,
  rightline=false,
  skipabove=10pt,
  skipbelow=10pt,
  leftline=false]{regbox}{Box}
  
\usepackage[p,osf]{cochineal}
\usepackage[T1]{fontenc}

\usepackage[scale=.95,type1]{cabin}
\usepackage[cochineal,bigdelims,cmintegrals,vvarbb]{newtxmath}
\usepackage[zerostyle=c,scaled=.94]{newtxtt}
\usepackage[cal=boondoxo]{mathalfa}
\usepackage{microtype}
\usepackage{array}
\usepackage{rotating}
\usepackage{comment}

\usepackage{etoolbox}
\apptocmd{\thebibliography}{\raggedright}{}{}

\usepackage{amsmath,amsfonts,bm}

\def\eqref#1{equation~\ref{#1}}

\DeclareMathAlphabet{\mathsfit}{\encodingdefault}{\sfdefault}{m}{sl}
\SetMathAlphabet{\mathsfit}{bold}{\encodingdefault}{\sfdefault}{bx}{n}

\title{Graph AI in Medicine
}

\author[1,2]{Ruth Johnson}
\author[1,3,$\ddag$]{Michelle M. Li}
\author[1,4,$\ddag$]{Ayush Noori}
\author[1,$\ddag$]{\\Owen Queen}
\author[1,5,6,7,$*$]{Marinka Zitnik}
\affil[1]{\small Department of Biomedical Informatics, Harvard Medical School, Boston, MA 02115, USA}
\affil[2]{Berkowitz Family Living Laboratory, Harvard Medical School, Boston, MA 02115, USA}
\affil[3]{Bioinformatics and Integrative Genomics Program, Harvard Medical School, Boston, MA 02115, USA}
\affil[4]{Harvard College, Cambridge, MA 02138, USA}
\affil[5]{Broad Institute of MIT and Harvard, Cambridge, MA 02142, USA}
\affil[6]{Harvard Data Science Initiative, Cambridge, MA 02138, USA}
\affil[7]{Kempner Institute for the Study of Natural and Artificial Intelligence, Harvard University, Allston, MA, USA\vspace{2mm}}
\affil[$\ddag$]{M.M.L, A.N. and O.Q. contributed equally (equal second authorship) and are listed alphabetically.\vspace{2mm}}
\affil[$*$]{Correspondence: \href{mailto:marinka@hms.harvard.edu}{marinka@hms.harvard.edu}}
\date{}

\begin{document}

\maketitle

\begin{abstract}
\noindent In clinical artificial intelligence (AI), graph representation learning, mainly through graph neural networks and graph transformer architectures, stands out for its capability to capture intricate relationships and structures within clinical datasets. With diverse data---from patient records to imaging---graph AI models process data holistically by viewing modalities and entities within them as nodes interconnected by their relationships. Graph AI facilitates model transfer across clinical tasks, enabling models to generalize across patient populations without additional parameters or minimal to no re-training. However, the importance of human-centered design and model interpretability in clinical decision-making cannot be overstated. Since graph AI models capture information through localized neural transformations defined on relational datasets, they offer both an opportunity and a challenge in elucidating model rationale. Knowledge graphs can enhance interpretability by aligning model-driven insights with medical knowledge. Emerging graph AI models integrate diverse data modalities through pre-training, facilitate interactive feedback loops, and foster human-AI collaboration, paving the way to clinically meaningful predictions.

\vspace{5mm}\noindent\textbf{Keywords:} Medicine, healthcare, graph neural networks, graph transformers, multimodal learning, transfer learning, human-centered AI, knowledge graphs
\end{abstract}

\section*{Introduction}\label{sec:intro}

With the simultaneous advancements of clinical information and omics-based technologies, the medical landscape has dramatically shifted towards a data-centric approach~\cite{council2011committee}.
Precision medicine aims to provide a more precise approach to disease diagnosis, treatment, and prevention based on individuals' unique medical histories and biological profiles~\cite{collins2015new}.  Clinical data, encompassing patients' medical history, patient-collected samples with genetic, genomic, and molecular information extracted from these samples, as well as data recorded during patients' interactions with the healthcare system, are poised to inform the development of powerful prediction models~\cite{safran2007toward}. 

However, healthcare systems optimize data collection and curation for patient care and healthcare administration instead of data-driven research. Most electronic health records (EHRs) come in structured, semi-structured, and unstructured data, including structured data tables, images, waveforms, and clinical notes. These datasets describe a multitude of complex and interconnected concepts, resulting in data that is both high-dimensional and heterogeneous in nature~\cite{data2016challenges}. However, EHRs often suffer from sub-optimal data quality (due to the fast-paced environment and lack of manual curation) and high levels of sparsity due to patterns of non-random missingness associated with healthcare practices~\cite{sandhu2012secondary, weiskopf2013defining}.  Due to these challenges related to the secondary use of EHR for precision medicine applications, transforming this information into a representation that computational algorithms can readily utilize is non-trivial and is currently an active area of study~\cite{jensen2012mining,hripcsak2013next}. 

Representation learning aims to automatically extract a computer-readable representation of the data, optimized in such a way that it captures valuable information for some downstream or target task (\textit{e.g.}, disease prediction)~\cite{bengio2013representation}. Most computational models assume input data is formatted in a grid-like structure such as a vector or matrix. For example, an image is a matrix with values representing the color at each pixel, and text can be tokenized into vectors. However, given the implicit connectivity of the human body, information regarding the relational structure underlying human biology should be fundamental when performing representation learning for clinical tasks. Enforcing a grid-like data format is also prohibitive to modeling multiple data modalities since inputs will not have a uniform shape across data types  (e.g., images, text, audio). However, given that decision-making performed by clinicians inherently relies on weighing multiple types of information--ranging from histopathology images to vital signs--limiting prediction algorithms to a single data modality would likely severely limit a model's performance and utility~\cite{acosta2022multimodal}. 

Graphs provide a principled way of explicitly modeling relational structure in data representations. These non-linear data structures present a formal framework for encoding complex relationships between objects or entities without imposing a strict ordering or shape. This highly versatile approach has been adopted in multiple biomedical applications, showing success in various clinical tasks (Table~\ref{tab:clinical_overview}). Graphs also balance flexibility and structure when representing data, allowing the natural integration of multiple data modalities and providing a strategy to leverage interdependencies across modalities. This relational structure can be solely defined by connections described in a patient's medical record history, and graphs can also contain links to external sources such as biomedical ontologies~\cite{himmelstein2015heterogeneous}. This enables the mathematical incorporation of prior domain-specific knowledge into guiding the patient representations~\cite{nelson2019integrating}. 
Not only are graphs naturally more interpretable to users, but algorithms operating on graphs have inherent algorithmic mechanisms for incorporating explainability, such as attention-based feature importance~\cite{vaswani2017attention}.

Here, we review graphs and graph representation learning in clinical AI. Although there are many innovative biomedical applications of graph representation in genomics~\cite{zhang2019hyper,wang2021scgnn,li2022graph}, proteins~\cite{cheng2008prediction, fout2017protein}, and therapeutics~\cite{zitnik2018modeling,jin2020modeling,zitnik2023current}, we focus on applications utilizing clinical data collected in the healthcare system setting. Many concepts and ideas described here apply across multiple application domains.  We begin with an overview of graph representation learning to highlight the key concepts relevant to this review but refer readers to prior works for a more in-depth review of graph representation learning theory~\cite{hamilton2017representation, hamilton2020graph, ju2023comprehensive}. We next discuss how graph representation techniques specifically address many of the current challenges in clinical AI algorithm development. Each of the following sections discusses the utility of graphs in transfer learning, multi-modal learning, and explainability.  We underscore the potential of graphs in clinical AI, emphasizing the convergence of multi-modal learning, transfer learning across patient populations and clinical tasks, and the imperative of human-centered AI.

\begin{small}
    \begin{table}

\begin{tabular}{p{8em} |p{20em}  |p{1em}  |p{1em}  |p{1em}  |p{1em}  |p{3em}} 

\textbf{Medical task} & \textbf{Description} & \textbf{\begin{turn}{-90}Relational data \end{turn}
} & \textbf{\begin{turn}{-90}Transfer learning\end{turn}} & \textbf{\begin{turn}{-90}Multimodality\end{turn}} & \textbf{\begin{turn}{-90}Explainability\end{turn} } & \textbf{\begin{turn}{-90}Example references\end{turn}} \\ \midrule

Drug repurposing and off-label prescription & The practice of identifying new therapeutic uses for existing drugs. Key challenges involve modeling the complex interactions between genes, pathways, targets, and drugs and the resulting exponential search space. & \ding{52} & \ding{52} & \ding{52} & \ding{52} & \cite{sosa2019literature, morselli2021network, nguyen2021graphdta, huang2023zero} \\ \midrule 

Medical safety and pharmacovigilance & Adverse drug reactions are the undesirable (potentially harmful) side effects associated with drug use. Key challenges include modeling complex interactions between biological and chemical processes and the effects of polypharmacy. & \ding{52} & \ding{52} & \ding{52} & \ding{52} & \cite{zitnik2018modeling, hwang2020drug}  \\ \midrule 

Imaging for neurological disease diagnosis & Techniques for the visualiation and assessment of brain function. A key challenge is incorporating various image modalities, including MRI, CT, PET, and EEG. & \ding{52} &  & \ding{52} &  & \cite{kim2013graph,tong2017multi,yang2019interpretable}  \\ \midrule 

Genetics for rare disease diagnosis & The characterization of medical conditions caused by extremely uncommon genetic mutations. The key challenge is the small size or complete lack of labeled cases. &  & \ding{52} & \ding{52} & \ding{52} &  \cite{li2019improving, sun2020disease, alsentzer2020subgraph, alsentzer2022deep} \\ \midrule

Interpretable prognostic biomarkers & The microscopic examination of tissues/cells to identify diseases or abnormalities in tissue samples. Whole-slide images are often broken into smaller images or patches due to size. A challenge is maintaining topological properties across sub-images. & \ding{52} &  &  & \ding{52} &  \cite{wu2022graph,lee2022derivation} \\ \midrule 

Disease-gene association for complex diseases & Inferring a relationship between specific genes/mutations and the susceptibility or development of a particular complex disease. Key challenges include modeling complex biological processes and the associated omics data. & \ding{52} & \ding{52} & \ding{52} & \ding{52} &  \cite{zhou2016knowledge, wang2019predicting, wen2023multimodal} \\ \midrule 

Early disease detection & Predicting the risk of an individual developing a disease based on their medical history as recorded in the EHR. Key challenges include the data quality and incompleteness associated with the secondary use of EHR. &  & \ding{52} & \ding{52} & \ding{52} &  \cite{nelson2019integrating, lu2021weighted,mao2022medgcn} \\ \bottomrule 
\end{tabular}
\caption{Example clinical tasks, descriptions, and how graph representation techniques can benefit. We also include a sample of references for each task; note that this list is not exhaustive.}
\label{tab:clinical_overview}
\end{table}

\end{small}


\section*{Algorithms for Learning  Graph Representations and Embeddings}\label{sec:definitions}

Graphs can encode rich relational structure in biomedical data, providing useful frameworks for representing EHR relational databases \cite{murali2023towards, walke2023importance}, medical ontologies \cite{gene2023gene}, as well connections between different therapeutics  \cite{chandak2023building}. Formally, a graph $\mathcal{G} = (\mathcal{V}, \mathcal{E})$ is defined by the sets of nodes (vertices) $\mathcal{V}$ and edges $\mathcal{E}$ with optional node features $\mathbf{X}^V$ and edge features $\mathbf{X}^E$. Traditional graphs are typically \textit{homogeneous} where nodes and edges are of a single type. We can also represent a graph where nodes and edges belong to multiple types, forming a \textit{heterogeneous} graph. For example, type 1 diabetes mellitus and insulin
can be represented as a disease and drug node in a graph with a shared edge between the two nodes denoting the disease as an indication for a specific drug.

We can leverage this relational structure encoded by graphs in various prediction tasks. First,  \textit{node prediction} refers to the task of learning a function ($f$) to predict either properties or labels ($y$) associated with a set of nodes ($v_i$) in a graph, $f(v_i) = y$. \textit{Edge prediction}, also known as \textit{link prediction}, aims to learn a function $f$ such that for any two nodes $v_i, v_j$, $f$ predicts whether or not an edge exists between the nodes, \textit{e.g.}, $f(v_i, v_j) = y \in \{ 0,1 \}$. \textit{ Graph prediction} seeks to learn a function $f$ that predicts a property $y$ describing an entire graph $\mathcal{G}_i$, e.g., $f(\mathcal{G}_i) = y$. 
Many more learning tasks can be performed on graphs, such as graph generation and node and graph clustering. For a comprehensive survey on graph learning tasks, please refer to Waikom et al. \cite{waikhom2023survey}.

\begin{figure}[t]
    \centering
    \includegraphics[width=0.9\linewidth]{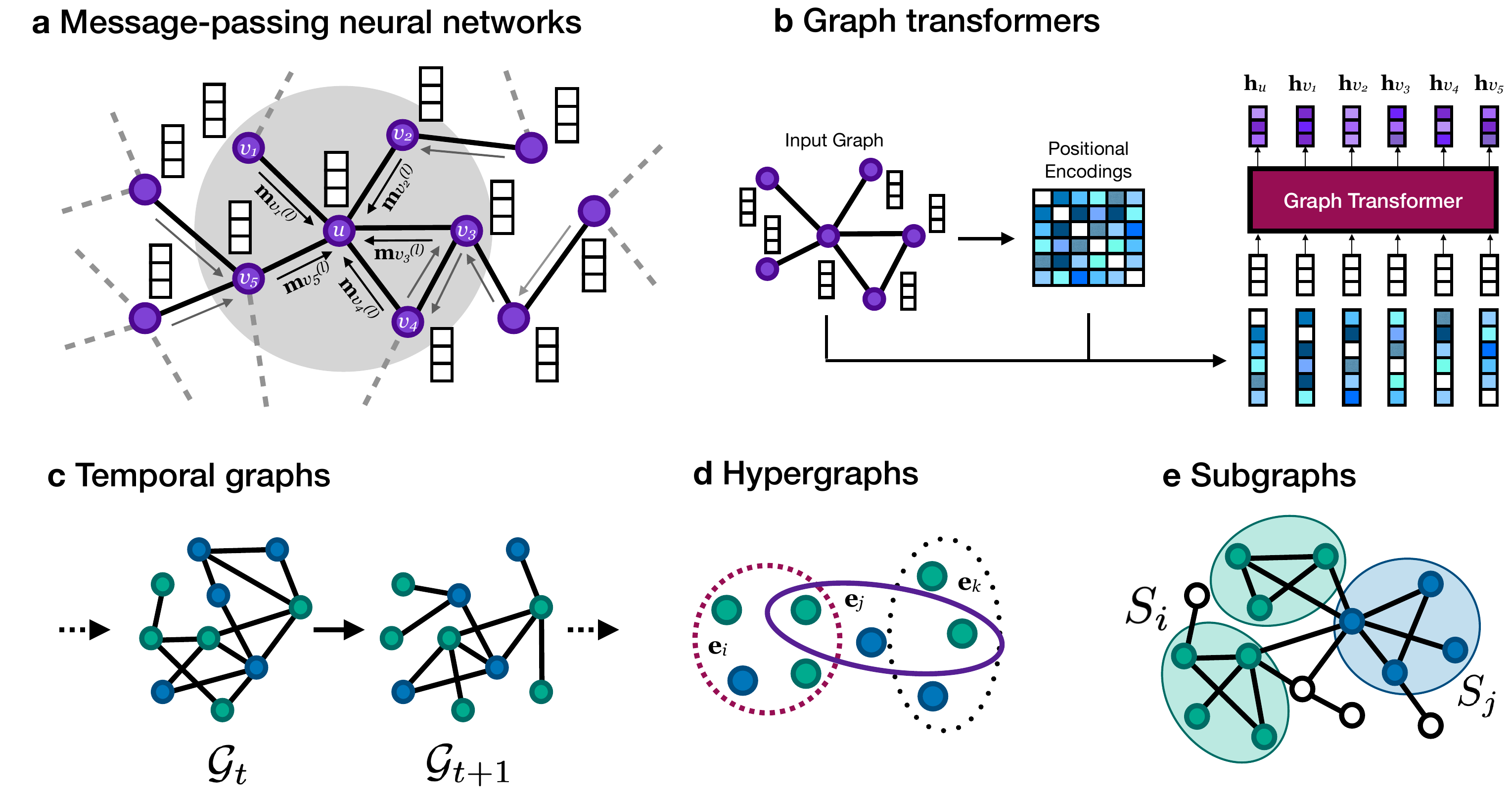}
    \caption{Variants of neural networks for graphs, including temporal graphs, hypergraphs, and subgraphs. a) shows a message-passing scheme by which messages are propagated along edges for each node. b) shows an example of a graph transformer, where the graph is expanded into a sequence with positional encodings. c) shows temporal graphs sampled over times $t$, $t+1$; these methods often use variants of a) or b) to learn representations at each step. d) shows a hypergraph, where hyperedges (rounded shapes) can encompass more than two nodes. e) shows a subgraph neural network, which learns representations for $S_i$, $S_j$, even when these subgraphs are disconnected.}
    \label{fig:arch}
\end{figure}

\subsection*{Representation Learning Basics}

Representation learning refers to learning an embedding space, \textit{i.e.} a space of vectors, that represents a given set of input data. 
We define a representation space or embedding space $z_1, ..., z_n \in \mathcal{Z}$, where each $z_i$ vector can represent nodes, edges, or entire graphs. 
Most often,  each $z_i$  represents a node, representing edges as a concatenation of $z_i$ and $z_j$. A graph can be represented by an aggregation of node embeddings where a given function, such as sum or average, is applied across the set of nodes in a graph.

Embedding methods are broadly grouped into those that produce shallow and deep representations. Methods that estimate shallow embeddings characterize nodes based on some predefined notion of distance within the graph, such as node co-occurrence encountered in random walks \cite{perozzi2014deepwalk}, local connectivity defined by graph traversal \cite{grover2016node2vec}, or relations defined within a knowledge graph \cite{trouillon2016complex, sun2019rotate}. 
The embedding space is then optimized to reflect this notion of distance. However, shallow embedding methods are inherently transductive where the learned embeddings do not generalize to nodes not observed in training. In contrast, deep embedding methods leverage neural network-based inference to learn a function that captures more complex relationships between nodes \cite{hamilton2017inductive}. This resulting optimized GNN allows for inductive learning where embeddings can be generated for nodes not present in the original training graph.

Embedding representations allow for the mathematical description of various prediction tasks.  For example, consider the task of finding patients with similar clinical profiles. Although a straightforward database query could be used, formally quantifying similarity among patients would be difficult to explicitly define. An embedding space naturally formulates this objective as a nearest-neighbor search. Given the embedding $z_i$ for a patient and some predefined distance function $D$ (\textit{e.g.}Euclidean distance), identifying similar patients involves finding the $m$ patients ${z_{j_1},...,z_{j_m}} \in \mathcal{Z}$ that have the smallest distance to $z_i$. This distance function gives a natural way to define patient similarity, avoiding costly expert-curated rules to determine a similarity metric.

\subsection*{Graph Neural Networks}

Graph neural networks (GNNs) are a class of neural network models designed for learning on graph-structured data. GNNs can learn representations of nodes, edges, and even entire graphs. The graph representation learning field has developed many different network architectures that each aim to capture different types of complex relationships within graph-structured data. For node-, edge-, and graph-level prediction tasks, message-passing and transformer-based architectures are among the most common.
 
\textbf{Message-passing neural networks (MPNNs)} are GNNs that consist of several layers that propagate information within the graph (Figure \ref{fig:arch}a). For a node $u$ at layer $l$, the MPNN gathers information from nodes around $v$ (called messages), aggregates these messages into a message feature $\mathbf{m}^{(l)}_u$, and then performs an update to the embedding $\mathbf{h}_u^{(l)}$ corresponding to node $u$. Mathematically, we can write this update as the following,
$\mathbf{h}_u^{(l)} = \textsc{UPD}(\mathbf{m}^{(l)}_u, \mathbf{h}_u^{(l-1)})$, where $\textsc{UPD}(\cdot)$ refers to a function that integrates messages with the current node embedding. Typically,  $\mathbf{h}_v^{(0)}$ is  initialized to node features for $u$, $X^V_u$. After propagating messages across several layers, the resulting  function is able to capture complex dependencies present within the graph. Variants of MPNNs modify the message-passing and update procedures \cite{hamilton2017inductive, xu2018powerful, xu2018representation, li2021training}, including the popular graph attention network (GAT), which uses an attention mechanism to learn along which edges to propagate information \cite{velickovic2018graph}.

\textbf{Graph transformers} are neural networks that directly feed node embeddings into a transformer architecture, omitting the use of message-passing. As depicted in Figure \ref{fig:arch}b, these models represent nodes as sequences and then positional encodings are appended to the input which incorporates relational information defined by the graph \cite{ma2023graph, kong2023goat}.  Graph transformers generally offer greater expressivity than MPNNs \cite{kreuzer2021rethinking}, and recent work has sought to combine MPNNs and graph transformers to increase model expressivity \cite{rampavsek2022recipe, wu2022nodeformer}.

\subsection*{Learning on More Complex Graphs} 

Graphs in healthcare may be dynamic over time or contain more complex relations between entities. Here, we highlight a few variants of GNNs that operate in challenging data regimes. 

\textbf{Temporal GNNs} learn on dynamic graphs, \textit{i.e.}, where nodes and edges change over some time steps. A dynamic graph $\mathcal{G}^D$ is defined as a sequence of graphs and associated time steps $1,..., T$, \textit{e.g.}, $\mathcal{G}^D = \{ \mathcal{G}_1, ..., \mathcal{G}_T\}$. Temporal GNNs learn representations for nodes at each time step, providing a formal mechanism for temporal predictions \cite{longa2023graph, skianis2023predicting, fritz2022combining}. These models can handle time-varying factors, such as patient states throughout their stay in the ICU~\cite{zhang2022graph}.

\textbf{Hypergraph GNNs} learn on hypergraphs, which are generalizations of graphs that allow for $n$-level connections between nodes through hyperedges. Message-passing is then performed along hyperedges instead of standard binary edges \cite{antelmi2023survey}. Hypergraph GNNs have been used to learn drug-disease relations \cite{pang2021hgdd} and EHR data \cite{sun2022ehr2hg, xu2023hypergraph, wu2023megacare}.

\textbf{Subgraph GNNs} learn representations for subgraphs, subsets of nodes, and edges in a larger graph. A subgraph is formally defined as $S = (V', E')$, where $V' \subseteq \mathcal{V}$ and $E' \subseteq \mathcal{E}$ for some graph $\mathcal{G} = (\mathcal{V}, \mathcal{E})$. A message-passing network operating on subgraphs learns a representation for $S_i$, \textit{i.e.}, $f(S_i) = z_i$ \cite{alsentzer2020subgraph, wang2021glass}. Subgraph GNNs help classify diseases, which can be represented as subgraphs of many phenotype nodes and edges~\cite{alsentzer2022deep}.


\section*{Inductive Biases and Transfer Learning on Knowledge Graphs} \label{sec:heterogeneous}

We proceed with a discussion of how medical knowledge can be integrated into neural network design to implement more reliable healthcare AI models. We highlight how medical knowledge, such as biological pathways and medical ontologies, can inform neural network structure through the use of relationships defined in biomedical knowledge bases. Not only can the incorporation of this domain-specific knowledge 
promote model interpretability but it also helps to avoid over-parameterizing deep learning models.

\subsection*{Designing Neural Architectures that Incorporate Medical Knowledge}

The underlying structure of heterogeneous graphs mirrors the connections that humans formulate when making decisions. For example, a biological pathway captures the series of interactions that must occur between molecules in a cell to create a gene product~\cite{hartman2023interpreting}. In a biologically and medically constrained neural network, the propagation of neural messages mirrors the logical transitions between ideas or concepts. As such, domain-specific graphs have been used to design the architectures of graph neural networks (Figure~\ref{domain}a).

Biological pathways provide a natural relational structure to inform neural network architectures. A recent work,
P-NET leverages hierarchical biological knowledge to integrate both broad and fine-grain biological processes into the model architecture. Propagating messages along the path of a biological pathway forces the model to leverage this existing knowledge to generate accurate and interpretable predictions~\cite{elmarakeby2021biologically, hartman2023interpreting}. Specifically, the input layer reads in the patient's profile, followed by three hidden layers of genes, pathways, and biological processes. The connections between P-NET's hidden layers are determined by known parent-child relationships between genes, pathways, and biological processes~\cite{elmarakeby2021biologically}.  Another recent work proposes using biologically informed neural networks (BINNs). The BINN architecture is formulated such that the input layer consists of proteins, the hidden layers are composed of biological pathways, and the output layer contains the biological process~\cite{hartman2023interpreting}. Training BINN with protein quantities as input enables biomarker identification and pathway analysis~\cite{hartman2023interpreting}.  By defining the network's layers and connections through biological relationships, biologically informed models avoid over-parameterization, provide enhanced interpretability, and incorporate external knowledge to improve predictions.

For clinical data and the associated medical concepts (\textit{e.g.}, diagnostic codes, lab tests, and medications), relational structure is often embedded through standardized medical ontologies~\cite{haendel2018classification}. A medical ontology is a structured representation (often hierarchical) of knowledge about medical concepts and their interrelationships~\cite{haendel2018classification}. These ontologies can provide a formal basis for incorporating clinical domain knowledge in prediction models. For example, hierarchical attention propagation (HAP) learns embeddings of medical concepts by hierarchically propagating attention weights across the entire medical ontology, outperforming prior approaches that only learn attention weights over ancestors~\cite{zhang2020hierarchical}. Other approaches leverage the hyperbolic space (\textit{e.g.} to effectively capture hierarchical structure~\cite{lu2019learning, hao2021medto}. Accurately capturing the granularity of medical concepts across the hierarchical levels through embeddings has been used to inform patient representations~\cite{peng2021mipo, yao2023ontology}, which can be represented as sets of embeddings corresponding to the medical concepts listed in their EHR.

Rather than aggregating embeddings of each node's direct neighbors, models may define a node's neighborhood based on the specific domain and task. For example, repurposing a drug entails identifying other diseases associated with the drug's biological target. So, finding these candidate new indications for the drug requires looking at nodes two hops away: a disease associated with the drug's target. This guilt-by-association principle has been incorporated directly into the model: for each drug or disease node, generate sequences of drug, protein, and disease nodes (using random walks) to learn semantic similarities between the nodes~\cite{bang2023biomedical}. As a result, the embeddings of drugs (or diseases) are near those of the proteins associated with the drugs (or diseases). New drug-disease relationships are inferred using embedding distance. Neighborhood aggregation strategies have also been designed to address a fundamental limitation in biology or medicine. Performing drug repurposing for poorly characterized diseases can be challenging due to the limited understanding of the affected genes and biological pathways. These query diseases' embeddings have been augmented by identifying similar diseases (using a domain-specific similarity metric) and aggregating them with the query disease~\cite{huang2023zero}. A similarity metric can be the number of shared pathways, for example. Augmenting the embeddings of these poorly characterized diseases has enabled zero-shot (\textit{i.e.}, no training examples) prediction of a drug's indications and contraindications~\cite{huang2023zero}.

\subsection*{Contextualizing Patient Data Using Medical Knowledge}

Patients are not treated in isolation, independently of any prior knowledge. They must be considered within the context of their medical history, existing knowledge about each disease or drug, and results of biological experiments. Such information can be encoded in a heterogeneous network. For example, patient-doctor encounters, including longitudinal hospital visits, may be modeled as a heterogeneous graph~\cite{biswal2020doctor2vec, hettige2019mathtt}. A knowledge graph, a particular type of heterogeneous graph, consists of interactions between drugs and proteins, associations between drugs, diseases, and proteins derived from clinical trials and population-level analyses, involvement of proteins in biological pathways, and more~\cite{fernandez2022integrating, chandak2023building, bang2023biomedical}. Medical ontologies and biomedical knowledge may also be combined into a single unified network~\cite{chandak2023building, wang2023stage}. There are many ways to consider patient information in specific contexts: overlaying patient features directly onto a reference graph, constructing patient-specific networks or subnetworks, and creating patient similarity networks via domain-specific similarity metrics (Figure~\ref{domain}b).

Overlaying patient data on a global reference graph effectively leverages its underlying topology to enrich predictions. It enables interpreting patient-level information by bridging healthcare process data and medical knowledge. Augmenting patient representations generated from medical records with representations learned from existing biomedical knowledge (\textit{e.g.}, by concatenating the embeddings generated for the two data modalities) is one way of integrating prior knowledge about diseases and drugs for diagnosing and treating patients in a more data-driven manner~\cite{gao2023medical}. Instead of integrating the independently generated representations, patient-specific data have been considered alongside the input graph. Multi-omic features of patients or diseases are directly incorporated in the network as node features~\cite{pfeifer2022gnn}. By predicting these node features during training, the model learns an integrated view of the diseases, thereby improving the identification of biomarkers, detection of disease modules, prediction of drug efficacy in patients, and interpretability of predictions~\cite{pfeifer2022gnn} For multimodal learning, where modality-specific models are used to integrate multimodal data, refer to the section, ``Multimodal Learning on Clinical Datasets.''

Patient data have also been used to alter the underlying graph or construct patient graphs/subgraphs. This strategy further distinguishes the learned patient representations, as their topologies differ depending on the included nodes and edges (\textit{i.e.},~based on their patient-specific data). Modifying the underlying graph, nodes that symbolize individual patients are added by connecting them to nodes in the graph corresponding to the patient's data (\textit{e.g.},~phenotypes, diseases, medications)~\cite{liu2020heterogeneous, li2022patient, bharadhwaj2021clep}. Features are also added to these edges based on the relevance of each node in the graph to the patient, such as high or low measurement relative to the distribution of values observed in the patient population~\cite{bharadhwaj2021clep}. Alternatively, edges are contracted to specify the topology tailored to a given disease in the underlying graph~\cite{pu2022integrated}. The graph representation learning model is trained on the new graph with patient nodes or disease-specific contracted edges. Rather than inserting nodes or edges, subgraphs of patients' data have been extracted from the underlying graph to train a subgraph-level representation~\cite{alsentzer2022deep}. This type of approach continues to optimize the representations of the underlying graph and that of the patient subgraphs. 

In contrast, personalized patient graphs are extracted and treated as independent graphs~\cite{wang2023stage}. For this use case, the underlying graph is no longer involved during training. For example, Stage-Aware Hierarchical Attention Relational Network combines a medical ontology with a biomedical knowledge graph, extracts personalized patient graphs from the underlying graph, simultaneously models the hierarchical structure of the ontologies as well as the heterogeneous relationships for each patient~\cite{wang2023stage}. Patient similarity is learned across these approaches that train on modified underlying networks or extracted patient graphs. To explicitly indicate that patients share similar patterns, patients can form a network where nodes are patients and edges indicate their similarity based on domain-specific criteria. Patient data and external biomedical knowledge have been used to construct patient similarity networks. The patient similarity is commonly based on the overlap of diagnostic codes (\textit{e.g.}, ICD) and basic features of the patients (\textit{e.g.}, age, sex)~\cite{tong2021predicting}. For rare disease diagnosis, accounting for the frequency of the causal genes (\textit{i.e.}, genes harboring the mutations causing the disease) and diseases in the similarity metric injects domain knowledge to and improves the predictive ability of the model~\cite{tong2021predicting}. GNNs can be used directly on patient similarity graphs or in combination with sequence models, such as recurrent neural networks, to predict future medical events.

\subsection*{Model pre-training on Patient Populations and Fine-tuning Across Clinical Tasks}

\begin{figure}[t]
\begin{center}
\includegraphics[width=0.75\textwidth]{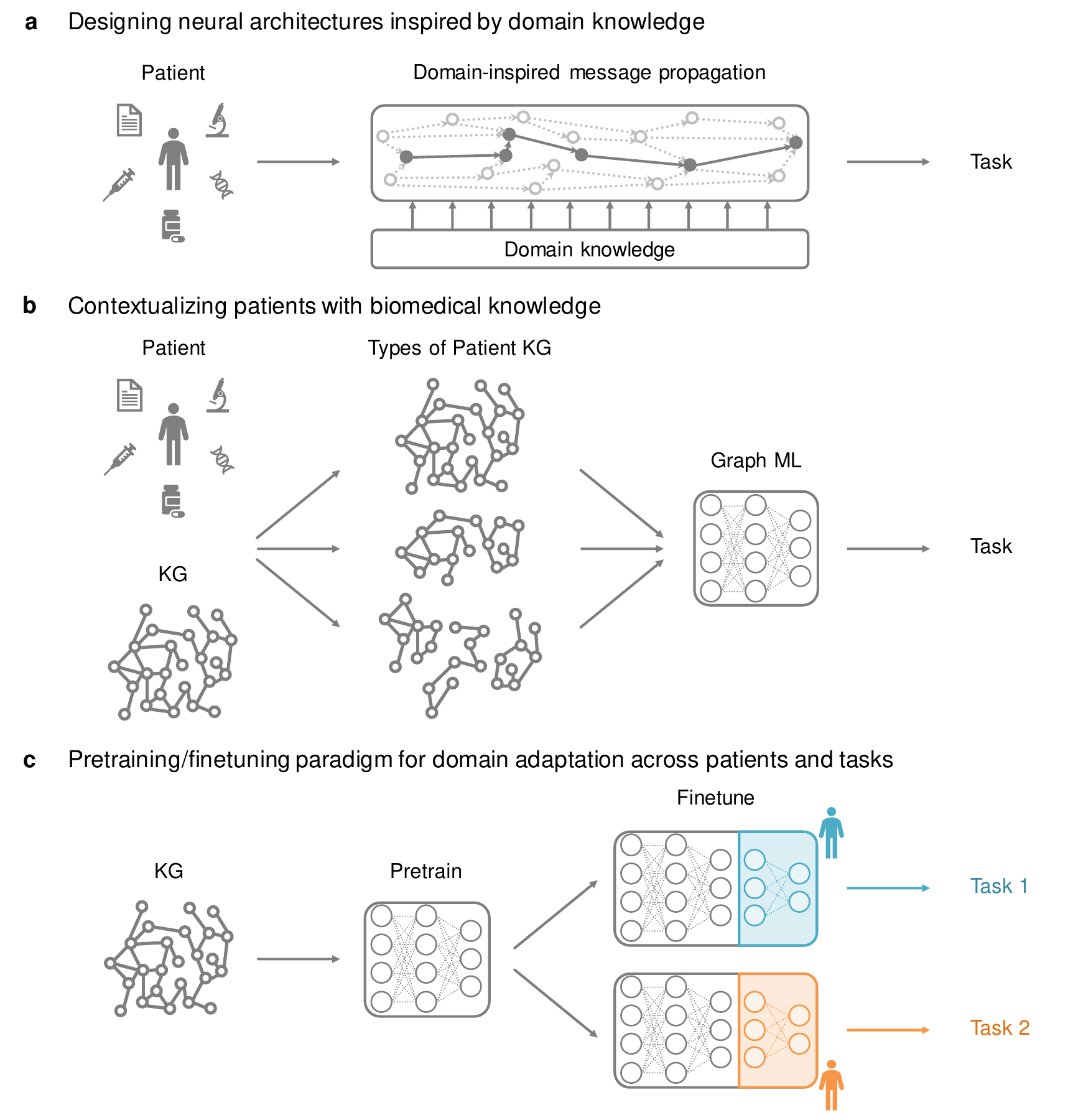}
\caption{Strategies for incorporating inductive biases and enabling transfer learning on knowledge graphs, including a) biologically informed neural network architectures, b) supplementing patient data with biomedical ontologies, and c) fine-tuning large pre-trained models across a broad range of clinical tasks.}
\label{domain}
\end{center}
\end{figure}

Annotated clinical datasets are often limited in size. However, traditional deep learning requires large annotated databases with millions of data points. As such, novel clinical AI algorithms must maximize the utility of the available data. The modern paradigm to do so is pre-training on a large-scale dataset without labels (\textit{i.e.}, unsupervised or self-supervised learning) and fine-tuning on a small labeled dataset for a specific task. Intuitively, the model learns implicit signals from a large-scale unlabeled dataset during pre-training and then transfers the knowledge to a specialized task while refining model parameters using a few labeled examples (Figure~\ref{domain}c).

The pre-training and fine-tuning paradigm for transfer learning has demonstrated broad success in clinical applications. Pre-training can be performed self-supervised on a general dataset unrelated to the medical domain~\cite{singhal2023large, singhal2023towards, kiyasseh2023vision}. To enable downstream medical Q\&A, models first learn patterns in language. MedPaLM~\cite{singhal2023large} and MedPaLM-22~\cite{singhal2023towards} are pre-trained on the general text and fine-tuned on medical questions and answers, which can be long-form or multiple choice. They demonstrate the benefit of instruction tuning, a data-, parameter-, and compute-efficient strategy to fine-tune the model for diverse clinical tasks. For clinical tasks on imaging, models are usually pre-trained on general images in a self-supervised manner. SAIS, a surgical AI system, leverages a visual transformer pre-trained on ImageNet via contrastive learning on augmented images to fine-tune for identifying intraoperative surgical activities~\cite{kiyasseh2023vision}. Pre-training has also been performed on large medical datasets. MedBERT~\cite{rasmy2021med} and NYUTron~\cite{jiang2023health} are pre-trained on electronic health records: MedBERT pre-trains on 28,490,650 patients via masked language modeling and predicting prolonged length of hospital stay. NYUtron pre-trains on 387,144 patients with 4.1 billion words through masked language modeling. Different data modalities have been combined for clinical applications. PLIP~\cite{huang2023visual} pre-trains on pathology images and captions by aligning their representations in a self-supervised manner and freezes the image decoder to fine-tune on text generation for specific clinical tasks. 

Graph-based approaches have also adopted the pre-training and fine-tuning paradigm for clinical applications. TxGNN pre-trains on a knowledge graph via link prediction and fine-tunes on predicting indication and contraindication relationships~\cite{huang2023zero}. SHEPHERD also pre-trains on a rare disease knowledge graph via link prediction but fine-tunes on a dataset of simulated rare disease patients~\cite{alsentzer2022deep}. Evaluated on rare disease patients in a zero-shot manner, SHEPHERD outperforms state-of-the-art approaches~\cite{alsentzer2022deep}. GODE pre-trains joint representation of drugs via contrastive learning based on their molecular graphs and knowledge subgraphs~\cite{jiang2023bilevel}. The resulting representations are fine-tuned using a multi-layer perceptron for a downstream task~\cite{jiang2023bilevel}. More recently, MolCap~\cite{wang2023molcap} pre-trains a graph transformer on four self-supervised tasks regarding molecular reactivity: predicting the formal charge, hydrogen content, chirality, and bond order between atom pairs that occur after a chemical reaction. Manual and automatically generated prompts are used to fine-tune the model for downstream tasks, such as drug toxicity and absorption~\cite{wang2023molcap}. G-BERT combines GNNs and text transformers to enable medication recommendation using structured electronic health records~\cite{shang2019pre}. G-BERT uses a graph attention neural network to generate diagnostic and medication code representations based on their neighborhood in the ICD-9 and ATC ontologies. During pre-training, G-BERT performs two tasks: predict the masked medical code in a single visit and predict the unknown drugs that treat a given disease or unknown diseases that are treated by a given drug. These approaches demonstrate the success of pre-training and fine-tuning on graphs, including molecular and knowledge graphs. However, other clinical data modalities complement networked data well.

Many approaches extract medical entities from clinical notes and map them to nodes in a knowledge graph. Using knowledge graphs to augment text can improve prediction stability, reliability, and interpretability while minimizing hallucination~\cite{pan2023unifying}. One approach involves aligning the medical concepts' representations generated from a knowledge graph (\textit{i.e.}, based on each concept's local neighborhood) and text (\textit{i.e.}, based on each concept's definition)~\cite{zhang2023knowledge}. Specifically for graph-based medical question (by doctor) and answering (by patient) system, the models either leverage pre-trained language models (on general or biomedical text) for continued pre-training on a publicly available medical question and answering dataset (\textit{e.g.},~MedDialog-EN)~\cite{varshney2023knowledge_aug, varshney2023knowledge} or perform contrastive learning to differentiate between medical entities during pre-training~\cite{xia2022lingyi}. In these approaches, knowledge graphs are used to generate complementary graph-based representations that are passed through the decoder alongside the language-based representations~\cite{varshney2023knowledge}, modify the masked ground truth token for token classification after the encoder layer (and the same hidden features are input to the decoder)~\cite{varshney2023knowledge_aug}, and narrow down symptoms to inquire about for identifying a disease~\cite{xia2022lingyi}. Refer to the section ``Multimodal Learning on Clinical Datasets'' for details about multimodal integration.

Looking forward, there are many opportunities to innovate. The utility of pre-training varies with domains and data modalities~\cite{mcdermott2023structure}; no one strategy is universally effective and beneficial across tasks~\cite{hu2019strategies}. Pre-training may not even be necessary~\cite{gao2023leveraging}. Further, prompting and instruction tuning are still not well-defined for graphs in general. However, prompting molecular graphs show that prompts can improve the downstream performance of GNNs~\cite{wang2023molcap}. 

\section*{Multimodal Learning on Clinical Datasets}\label{sec:multimodal}
In clinical datasets, structured and unstructured features often describe patients across various modalities \cite{acosta_multimodal_2022}. For example, EHRs may encompass free clinical text, family history, imaging and laboratory studies, medications, and billing codes. Additionally, biobanked patients may have matched --omics profiling (\textit{e.g.}, whole-exome sequencing). External biodata available from wearable biosensors and other external data sources can provide measurements of environmental exposures that may impact patient health. Integration of these complementary modalities promises to offer a unified view of patient health, enabling computational models to make precise and personalized clinical predictions.
Fusion of multimodal clinical data can occur at different stages in the modeling process, allowing information learned from each data channel to inform other modalities. These fusion strategies can be delineated into three broad categories: early, intermediate, and late integration \cite{lipkova_artificial_2022, kline_multimodal_2022}.

\subsection*{Early Integration}

In early integration, fusion occurs at the data level. Diverse data sources are transformed into the same feature space and passed as input to an unimodal architecture. The interactions between biomedical entities can be explicitly modeled in a graph-structured format such as a heterogeneous knowledge graph, where knowledge-informed descriptions and relationships are encoded into graphs upon which a GNN can learn \cite{chandak2023building, ektefaie_multimodal_2023}. For example, in LIGHTED, multimodal clinical features from patient encounters are extracted via manually defined feature engineering steps to form patient-feature-encounter graphs \cite{dong_integrated_2023}. A heterogeneous relational GNN embeds these multimodal graphs; raw feature vectors and encounter node embeddings are provided to a long short-term memory (LSTM) network to learn patient trajectories for opioid overdose prediction. 

Other examples of early integration include genomic data integration, where genomic datasets, like gene expression profiles, DNA methylation, and single nucleotide polymorphisms (SNPs), are integrated at the data level. These diverse datasets can be represented as multi-layered graphs, where each layer corresponds to a specific genomic data type. GNNs or graph transformers can be trained on these integrated graph structures to predict phenotypic outcomes or disease susceptibilities. In another example, brain imaging data from different modalities, such as fMRI, MRI, and DTI, can be integrated early on. After integration, this data can be represented as brain connectivity graphs where nodes represent brain regions and edges represent connections or activations. GNNs are then applied to these graphs for tasks like disease classification or cognitive state prediction. However, by lacking modality-specific encoders that can learn the idiosyncracies of each modality, early integration strategies may not fully exploit cross-modal interactions.

\subsection*{Intermediate Integration}

In intermediate integration, fusion occurs at the encoder level. Unlike early integration, modality-specific encoders are combined in the model to create unified representations; unlike late integration, information from different modalities is combined during convolution rather than after. Loss from the objective function propagates back to modality-specific channels whose outputs are often successively aggregated to create unified embeddings. Various clinical features, events, and outcomes have been successfully predicted using intermediate integration approaches on graphs, including patient survival in cancer \cite{hou_hybrid_2023}; adverse drug events \cite{krix_multigml_2023}; freezing of gait from footstep pressure maps and video recordings in Parkinson's disease \cite{hu_graph_2023}; and steatosis, ballooning, and fibrosis in histological stains of nonalcoholic steatohepatitis \cite{dwivedi_multi_2022}.

Intermediate integration offers the most flexibility to design integration strategies informed by prior knowledge of how different patient-level datasets depend on each other. For example, the Knowledge-enhanced Auto Diagnosis (KAD) model uses three tiers of integrated learning to combine a medical knowledge graph with radiology images and correlated textual descriptions for the automated diagnosis of chest X-ray images \cite{zhang_knowledge-enhanced_2023}. 
During training, radiology images are passed to an image encoder (\textit{e.g.}, ResNet-50 or ViT-16) while the matched textual descriptions are passed to a language encoder; the outputs of each encoder are then jointly passed through a transformer architecture to provide a final prediction.

Graphs can also enable domain knowledge-informed modality integration. For example, in studies of the human brain, graph-based integration of magnetic resonance imaging and diffusion tensor imaging (DTI) modalities has been informed by expert-created brain anatomical atlases, such as the Automatic Anatomical Labeling (AAL) atlas, with AAL Regions of Interest (ROIs) as nodes \cite{dsouza_m-gcn_2021, cai_graph_2023}. In a recent study\cite{cai_graph_2023}, modality-specific convolutional autoencoders learn representations of structural MRI and DTI scans. An AAL-informed brain network fuses the autoencoder outputs; a graph transformer then learns this network to estimate brain age as a biomarker for Alzheimer's disease. Together with KAD, these examples demonstrate how graph-guided learning can integrative structural and functional connectomics data at the representation learning and integration level to provide multidimensional phenotypic characterization. 

\subsection*{Late Integration}

Finally, in late integration, fusion occurs at the output level. Different unimodal architectures are trained individually or jointly to learn modality-specific representations, which can then be aggregated for downstream prediction. Graph-based late integration strategies have allowed the prediction of 30-day-all cause readmission \cite{tang_predicting_2023}, survival analysis in cancer \cite{azher_development_2022, gao_predicting_2022}, and diagnosis of mild cognitive impairment \cite{liu_mmgk_2022}, among other applications. In these architectures, GNN can play either role: the feature extractor or the feature aggregator. For example, in MOGONET, weighted sample similarity networks are constructed from mRNA expression data, DNA methylation data, and microRNA expression data; these networks are then used to train -omics-specific GNNs, whose predictions are combined in a fully connected final network for patient classification \cite{wang_mogonet_2021}. Both the unimodal encoders and the final ``View Correlation Discovery Network'' are jointly trained. In MOGONET, the GNN acts as the unimodal encoder but can also act as the feature integrator. For example, in DeepNote-GNN, a pre-trained BERT language model is used to generate embeddings from clinical admission notes; a GNN then learns on the similarity graph from these representations for prediction of 30-day hospital readmission \cite{golmaei_deepnote-gnn_2021}. 

Within the medical realm, integration techniques—encompassing early, intermediate, and late stages—have established a critical foundation for harnessing information from varied sources and modalities. Despite its innovative utilization of graph-based configurations, early integration encounters limitations, primarily stemming from the lack of encoders tailored to individual modalities. By contrast, intermediate integration offers a sophisticated equilibrium. Late integration adopts a more phased approach, permitting comprehensive learning specific to each modality before fusion. Moving forward, these integration paradigms highlight the necessity of developing methods based on data characteristics and clinical use cases, including considering end users and existing workflows within which the methods are implemented. 

\begin{figure}[t]
\begin{center}
\includegraphics[width=0.65\textwidth]{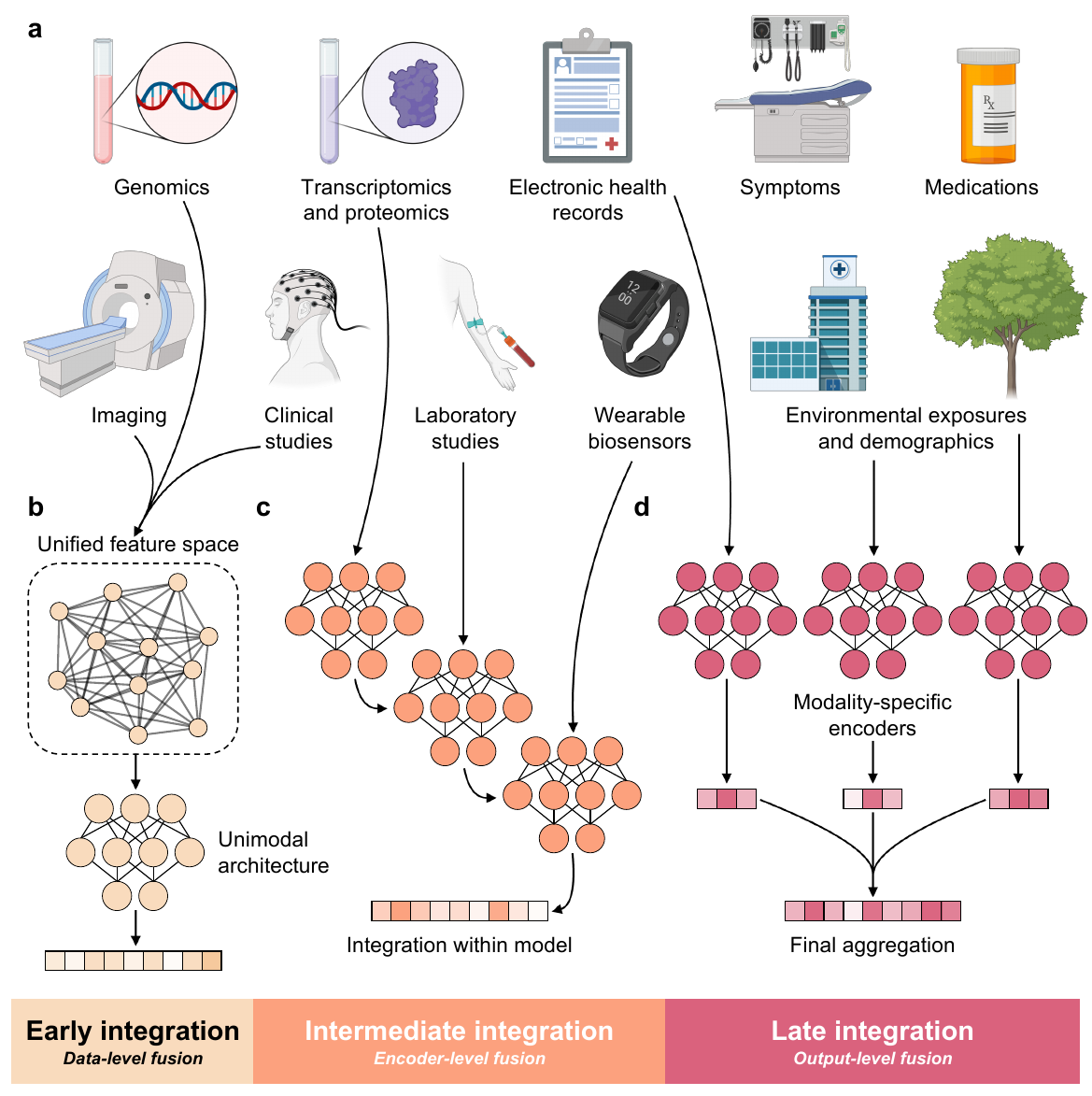}
\caption{Deep learning on a) multimodal clinical datasets including omics-data, medical imaging, EHR, and more. Data modalities can be fused through b) early, c) intermediate, and d) late integration strategies. Elements of this figure were created using BioRender.}
\label{multimodal}
\end{center}
\end{figure}

\section*{Explanability with GNNs} \label{sec:interpretability}

Although the accuracy of clinical AI algorithms has achieved remarkable prediction performance, usability requires building trust within the clinical community. Providing model transparency and explainability are crucial to attaining domain experts' trust and confidence in these models for high-stakes decision-making \cite{agarwal2022probing, bienefeld2023solving}. Model \textit{explainability} broadly describes the degree to which a model and its predictions can be communicated through clear, logical descriptions. These descriptions can be based on visualizations, text, statistics, or counterfactuals but do not necessarily assume that the user knows the exact model construction details \cite{agarwal2023evaluating}. 

In practice, the processes and methodologies used to achieve each concept depend on the relevant stakeholders and the context of its downstream use. Stakeholders are defined as anyone who interacts with or is affected by the decisions made based on the model's output \cite{bhatt2020explainable}. Providing model explanations for clinical AI models is unique since the resulting decisions propagate to diverse users with various needs and backgrounds, including clinicians, nurses, technicians, patients, and policymakers~\cite{davenport2019potential}. Meeting the needs of the full spectrum of stakeholders requires solutions that leverage different explanation modalities (Figure \ref{sec5fig}). The flexibility of graph representation learning algorithms provides a unique basis for constructing model explanations that can be tailored to different stakeholder audiences. Most explainability techniques for deep graph representations are local (sample-level) explainability~\cite{yuan2022explainability}. 

\subsection*{Mathematical Explainability}

Traditional methods rooted in the explainable AI machine learning field typically rely on \textit{mathematical}  modalities of explanation \cite{yang2022unbox}. These techniques tend to be more statistical- or data-oriented and focus on the internal mathematical mechanisms of a model. Often, these forms of explanation require a degree of a priori understanding or knowledge of the prediction model and problem setup. Explanation techniques vary according to where they are incorporated into the modeling process. \textit{Ante hoc} techniques involve incorporating mechanisms for explainability directly within the model from its initial construction. The parameters describing the predictor's explanation are learned during the training process \cite{henderson2021improving, miao2022interpretable}. For example, the attention mechanisms used in graph attention networks are learnable weights indicating the amount of contribution from a given neighbor node. In contrast, \textit{post hoc} explainability techniques are only applied after the regular training process is completed, thus not altering the original prediction model. One of the most widely used post-hoc explanation methods is SHAP \cite{lundberg2017unified}, which leverages concepts from game theory to estimate feature contribution scores. This has also been adapted for graph neural networks where contribution scores are contributed on nodes and their features \cite{duval2021graphsvx}.

Explanations can be presented in various forms or modalities. The first mathematical modality is based on attribution maps, which give importance scores over the features in a graph; most often, these scores are given over graph nodes \cite{pope2019explainability}, edges \cite{schlichtkrull2020interpreting}, or both \cite{ying2019gnnexplainer}. Attribution scores can also be presented as discrete values, such as $\{0,1\}$, which has the effect of selecting discrete entities in a graph, such as a subgraph or set of nodes \cite{yuan2021explainability, schlichtkrull2020interpreting}, rather than continuous scores over every candidate node in the graph \cite{pope2019explainability}. In previous literature, attribution maps are the most explored modality for method development. A second modality of explanations in graphs is that of the counterfactual explanations. These explainers present information in terms of counterfactuals, i.e., the assessment of alternative scenarios, such as predicting another label for a given node. The critical insight of counterfactual explainers is that interpreting multiple counterfactuals for one sample, as opposed to one explanation via an attribution map, increases human interpretability. Counterfactual explainers for graphs often focus on providing some perturbations to graph structure that might cause a change in prediction \cite{ma2022clear, lucic2022cf}. Such explanations may be necessary for healthcare, where assessing alternative outcomes for predictions, such as from diagnostic systems, dramatically increases the transparency for downstream users such as clinicians.

\subsection*{Human-Centered Explainability and Interactions}

There is often a divide between what constitutes an optimal explanation for clinicians versus data scientists or machine learning researchers. For example, developers use explainable AI tools to debug, probe, and audit models, whereas researchers utilize these tools to discover novel patterns in data \cite{bienefeld2023solving}. Clinicians use explainable AI as a single tool in their overall assessment of a patient, including their current patient observations and knowledge from clinical practice \cite{henry2022human}. Unlike pure discovery, results must fit within the current clinical context and known medical literature -- thus, an explainability tool in this setting would need to demonstrate that it satisfies these criteria for it to be trustworthy to clinicians \cite{schwartz2022factors}.

Given the natural logic structure imposed by graphs, graph representation models offer inherent opportunities for user-focused explainability. Prior studies have constructed interactive user interfaces for counterfactual queries regarding a graph's edges and nodes 
\cite{beinecke2022interactive}. Additionally, a prior work conducted a formal user study for drug repurposing prediction where path-based explanations can be analyzed and compared~\cite{wang2022extending}. The overall positive feedback from medical professionals described in the user study demonstrates graph-based user explanations as a promising technique for biomedical explanations.

\begin{figure}[t]
\begin{center}
\includegraphics[keepaspectratio=true,scale=0.35]{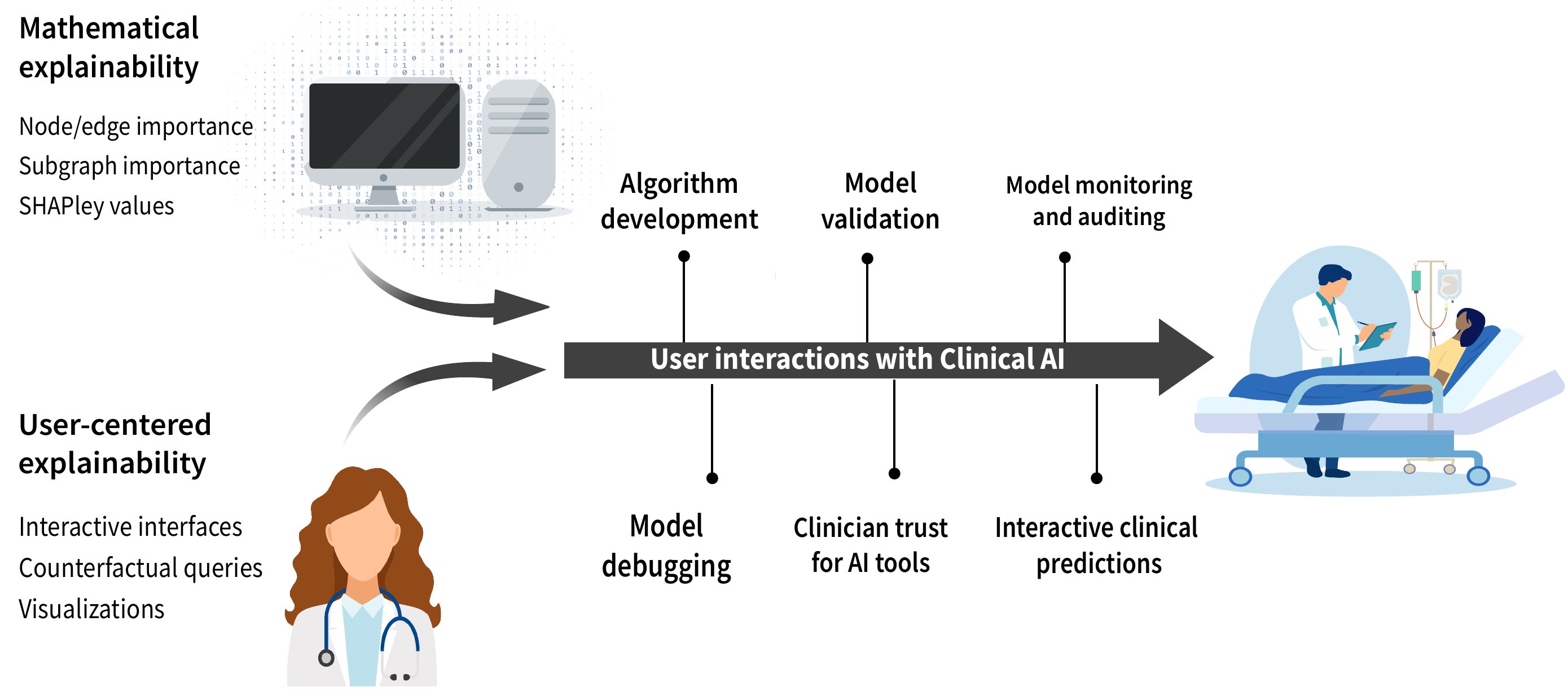}
\caption{Mathematical and user-centered explainability for clinical AI tools affect various stakeholders across data scientists, researchers, clinicians, patients, and administrators.}
\label{sec5fig}
\end{center}\end{figure}


\section*{Challenges and Future Directions} \label{sec:future}

While graph AI holds tremendous promise for advancing biomedical research and healthcare, numerous challenges and practical considerations still need to be addressed to fully utilize these techniques' potential. A key benefit of graph models is the ability to naturally incorporate multi-modal datasets such as genetic sequences, biomarker measurements, imaging data, and more through heterogeneous graph models. As these datasets become more complex, fusing this information into a cohesive graph representation will require advanced data harmonization, feature engineering, and graph construction techniques. An inherent challenge of modeling multiple modalities is the issue of \textit{missing modalities}, which is akin to missing data, except the missingness is by modality type~\cite{ektefaie2023multimodal}. Accurately modeling these patterns is especially relevant in the healthcare setting where we cannot assume that all types of measurements will be uniformly collected over individuals~\cite{zhang_m3care_2022} and much of the missingness in clinical data is non-random and may even be informative for certain problems~\cite{groenwold2020informative}. 

Moreover, clinical data modalities inherently have very intricate relational dependencies. For example, a clinical marker may be reflected in a laboratory test, clinical note, or radiology report. Still, depending on the exact modality, the record may have varying contexts and interpretations for diagnosis. Complex relations between modalities, such as this, can lead to \textit{modality collapse} where in the learning process, the model solely focuses on a subset of the provided modalities~\cite{ektefaie2023multimodal}. This issue can occur when specific modalities are beneficial during the training process and others are ignored. However, those modalities may be informative when the model is asked to make new predictions during inference. The study of modality collapse in general and specific techniques for addressing this phenomenon when modeling biomedical data is an active area of ongoing research.

The scalability of graph neural networks remains a practical concern, mainly when dealing with large-scale, high-dimensional biomedical datasets. Scaling computations involves exploring distributed computing, parallel processing, and hardware acceleration techniques to handle the computational demands of training complex graph models on extensive biomedical datasets. For example, modeling long-range connections in a large graph would require numerous layers (a $k$-layer GNN considers neighbors at most $k$-hops away), but most GNNs are only a few layers deep due to computational constraints. However, increased scalability is not solely dependent on increased hardware capabilities. Even methods that attempt to scale past $k=3$ layers run into issues of over-smoothing where node representations of different classes are indistinguishable, which is hypothesized to be a result of the message-passing scheme.

Although mini-batching is a common technique for modeling datasets too large to fit on a single GPU, creating random graph partitions does not always ensure effective training. This straightforward approach can lead to batches with highly imbalanced node and edge types when modeling heterogeneous graphs, prompting a new heterogeneous mini-batch sampling algorithm to address these concerns. Thus, developing algorithmic techniques that can efficiently learn representations from massive graphs will also be essential to applying graph-based methods to high-dimensional medical datasets. 

Additionally, ensuring the interpretability and reliability of graph-based models is crucial in biomedical applications to ensure clinical AI algorithms are fair and do not exhibit biases, which can contribute to or exacerbate health inequities. Researchers must develop techniques that optimize prediction performance and provide insights into the learned representations and decision-making processes, enabling clinicians and researchers to validate and understand a model's outcomes. Most graph representation methods that explicitly handle issues of fairness directly borrow techniques (as well as inductive biases) from non-graph domains. However, these approaches do not consider how sensitive information and associated biases can be propagated through the network during feature propagation. Even after a model has been developed, researchers must be vigilant about model auditing and post-market surveillance. Because medical data is recorded in a dynamic, non-stationary environment, dataset shifts constitute a major concern because these distributional changes can lead to unexpected biases or errors after a model has been deployed. Given the multiple modalities often used in graph AI models, monitoring dataset shifts and the interaction between the shifts across modalities is even more complex.

The fast-evolving landscapes of graph representation learning and digital health present exciting future opportunities in clinical AI. Clinical prediction tasks are inherently complex due to the nature of the tasks and only a few (if any) accompanying labeled data points from each task. However, with the advancement of ultra-large prediction models with billions of parameters, machine learning is shifting from single-task-oriented models to generalist models that can solve diverse tasks. The approach of constructing models trained on broad datasets and later adapted for more specialized tasks is referred to as \textit{foundation models}~\cite{bommasani2021opportunities}. A key factor to this approach is using unlabeled data and then leveraging techniques in self-supervised or semi-supervised learning. This eliminates the need for large amounts of labeled data, which is particularly expensive and time-consuming to collect in the medical domain. 

Foundation models naturally fit within the pre-training and fine-tuning paradigm, where a base model is first pre-trained on a large, broad dataset. Then, typically, a subset of the based model parameters is updated to adapt the model to an array of clinical tasks. Graphs naturally allow the integration of multiple data modalities and fusion techniques through heterogeneous graph representations~\cite{mcdermott2023structure}. Additionally, knowledge graph-based foundation models could infuse medical knowledge in pre-trained models, enabling models to perform predictions in a few-shot or zero-shot manner with no or minimal fine-tuning. As the field moves toward constructing these large, multi-purpose models, techniques for learning more expressive, explainable, and compelling representations will continue to develop.

\newpage

\section*{Acknowledgements}

R.J., M.M.L., A.N., O.Q., and M.Z.~gratefully acknowledge the support of NIH R01-HD108794, US DoD FA8702-15-D-0001, awards from Harvard Data Science Initiative, Amazon Faculty Research, Google Research Scholar Program, AstraZeneca Research, Roche Alliance with Distinguished Scientists, Pfizer Research, Sanofi iDEA-iTECH Award, Chan Zuckerberg Initiative, and Kempner Institute for the Study of Natural and Artificial Intelligence at Harvard University. A.N. was supported, in part, by the Summer Institute in Biomedical Informatics at Harvard Medical School. M.M.L.~is supported by T32HG002295 from the National Human Genome Research Institute and a National Science Foundation Graduate Research Fellowship. R.J.~is supported by the Berkowitz Family Living Laboratory at Harvard Medical School. Any opinions, findings, conclusions or recommendations expressed in this material are those of the authors and do not necessarily reflect the views of the funders.

Figure \ref{multimodal} was created with BioRender.com under agreement number HY25Y3F9Q0. Images utilized in Figure \ref{sec5fig} were obtained under an Adobe Stock Education License:
\textcopyright ink drop/Adobe Stock, \textcopyright ssstocker/Adobe Stock, \textcopyright Feodora/Adobe Stock, \textcopyright santima.studio (02)/Adobe Stock.

\bibliographystyle{ar-style3}
\bibliography{references}

\end{document}